\documentclass[peerreview]{IEEEtran}
\usepackage{cite} 
\usepackage{url} 
\usepackage[utf8]{inputenc} 
\usepackage{booktabs} 
\usepackage{graphicx}  
\usepackage[T1]{fontenc} 
\usepackage{textcomp}
\usepackage{ae} 
\usepackage{amsmath}
\usepackage{amsfonts}
\usepackage{amssymb}
\usepackage{array}
\usepackage{blindtext}
\usepackage{dcolumn}
\usepackage{verbatim}
\usepackage{enumitem}
\usepackage{parskip}
\usepackage[table,xcdraw]{xcolor}
\usepackage{lipsum}
\usepackage{pifont}
\usepackage{afterpage}
\usepackage{imakeidx}
\usepackage{setspace}
\usepackage{pbox}
\usepackage{wrapfig}
\usepackage{dsfont} 
\usepackage{ulem}
\usepackage{bbm}
\usepackage{MnSymbol}
\usepackage{color}
\usepackage{makecell}
\usepackage{footnote}
\usepackage{tablefootnote}
\usepackage[numbers]{natbib}

\newcounter{rowno}

\begin{document}
\title{Theoretical Foundations of the A2RD Project: Part I}


\author{Juliao Braga$^{1,2}$, Joao Nuno Silva$^2$, Patricia Takako Endo$^{3,4}$, Nizam Omar$^1$ \\
$^1$Universidade Presbiteriana Mackenzie, BR\\
$^2$IST - INESC ID, University of Lisboa, PT \\
$^3$Universidade de Pernambuco, BR\\
$^4$Dublin City University, IE\\
}
\date{28/08/2018}

\maketitle
\tableofcontents
\listoffigures
\listoftables

\IEEEpeerreviewmaketitle
\begin{abstract}
This article identifies and discusses the theoretical foundations that were considered in the design of the A2RD model. In addition to the points considered, references are made to the studies available and considered in the approach.
\end{abstract}


\section{Introduction}

In 2014 began the first movements characterizing the interests of authors in the application of Artificial Intelligence in the Internet Infrastructure \cite{Braga:2014a}. The notion of intelligent agents \cite{Russel:2010}, was already perceived as feasible to be applied in the various areas of knowledge, in particular, those that affect the resources and facilities that make the Internet work.

To be more precise, the interests were to apply intelligent agent techniques, now called Intelligent Elements (IE) in restricted domains of the Internet Infrastructure, that is, in the so-called routing domains represented by Autonomous Systems (AS) \cite{Braga:2018b}. Many terms used or need to be used to form the set of concepts necessary to apply the specific notion of IEs are still not well defined or clear in the literature. To avoid ambiguities and therefore to leave the understanding crystalline, it becomes necessary some definitions that will be used in the text.

This paper deals with these definitions and concepts, as well as the theoretical framework that guarantees the perfect understanding of the possible applications of model A2RD on the Internet Infrastructure. In addition to the Introduction, the paper has six more sections clarifying the notions of: \textit{Self-organization}, \textit{Domains}, \textit{Interoperability}, \textit{Intelligent Agents}, \textit{Multi-Agents} and \textit{Communication between Intelligent Agents}.

\section{Self-organization}

\begin{figure}[!ht]
\centering
\includegraphics[width=1\columnwidth]{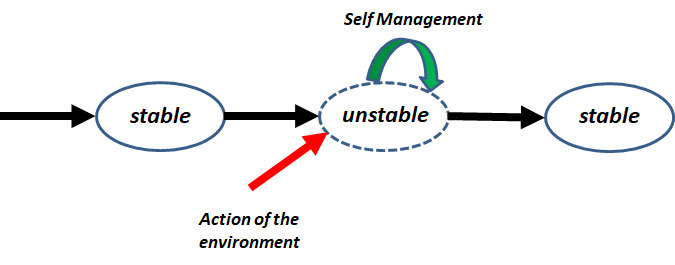}
\caption{Self-organization. Adapted from \cite{Braga:2015}}
\label{fig:autoorganization}
\end{figure}

The notion of \textit{self-organization} was awakened with a certain intensity in the early 1970s by Ilya Prigogine, Nobel Prize in Chemistry in 1977, in relations with natural systems \cite{Ebeling:2011}. In the context of this work, when dealing with intelligent agents, by \textbf{self-organization} is meant the ability of a agent to react readily when it perceives in some way a threat (or instability) in its environment indicating the possibility of a deviation in their functional objectives. This reaction returns the functionality of the agent to its stable condition existing in the previous state (self-organized) to the extraordinary event of the environment. The Figure \ref{fig:autoorganization} is a simplified view of this definition. 

\begin{figure}[!ht]
\centering
\includegraphics[width=1\columnwidth]{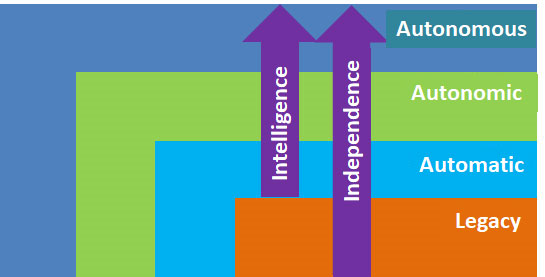}
\caption{Characterization of intelligence and independence of Intelligent Systems in the A2RD project}
\label{fig:emergence}
\end{figure}

For a \textbf{intelligent agent} or \textbf{intelligent system} or \textbf{Intelligent Element} (IE) to be able to self-organize, it must have certain properties and appropriate functional characteristics, which will be seen later. For the purpose of this text, if an \textbf{intelligent system} exercises its capacity for self-organization without absolutely no human intervention, it will be recognized as \textbf{autonomous}, with definite freedom to act \cite{Agoulmine:2010}. If, however, to exercise the function of self-organization, the intelligent system depends on a human orientation, not directly, but through pre-defined parameters indicating how to react, then the given denomination will be \textbf{autonomic}. Put differently, a \textbf{intelligent system} is \textbf{autonomic} if there is a human orientation (a plan) about how it should behave to self-organize, when reacting to an action of the environment. If an \textbf{intelligent system}, in order to exercise its capacity to self-organize, performs functions integrally put by the human being, then it is considered \textbf{automático}. Finally, if the intelligent system has not the conditions to self-organize, that is, under an unusual action of the environment, it eventually fails, then it is said to be \textbf{legacy}. The Figure \ref{fig:emergence} shows the relationship between such \textbf{intelligent systems} indicating the degrees of \textbf{independence} and \textbf{intelligence} aggregates and how it will be interpreted in this text.

The notion of \textbf{independence} is linked to human participation. The more intense that participation, the less autonomous the system is. Already, the notion of \textbf{intelligence} is associated with the ability of \textbf{self-organization} of the system. The greater the ability to effectively use the properties and functionality of self-organization, the more autonomous the system is. On the other hand, the notions of intelligence and independence are directly associated with the learning ability of the intelligent elements. This learning will be effective if there is \textbf{cooperation} between the intelligent elements, without distinction. The project is not interested in the functional aspects of \textbf{legacy systems} but, whenever necessary, will use results or configuration parameters of these systems, in order to maintain balance between all the elements involved, in the pursuit of their goals.

\section{Domains}

The word domain represents a collection of things (actors, entities, etc.) that are aligned and united around common goals, within the specific limits of a particular area of interest\footnote{\url{http://www.ncoic.org/cross-domain-interoperability}}. In the context of the A2RD project, the area of interest is the environment outlined by the activities associated with the ASs, which together represent the Internet.

ASs, in turn, have subsets of interests that are identified as \textbf{subdomains}.

\section{Interoperability and Ontology}

\textbf{Interoperability} is the term used in this context to designate the ability to cooperate between domains and / or subdomains through their respective IEs, in the achievement of common objectives. There are two types of interoperabilities: \textbf{syntactic interoperability} and \textbf{semantic interoperability}.

\textbf{Syntactic interoperability} refers to information that is exchanged between IEs during the connection. Such information, in general, is inserted in the context of the protocols that allow such connectivity. Eventually, additional information, which ascends to the upper layers of the TCP / IP model, uses additional information to maintain the connection that does not need interpretation of meaning, or are just pure data. In this case, some features such as \textbf{eXtensible Markup Language} (XML) or \textbf{Structured Query Language} (SQL) in formats recommended and described in this document enable interoperability.

\begin{figure}[!ht]
\centering
\includegraphics[width=1\columnwidth]{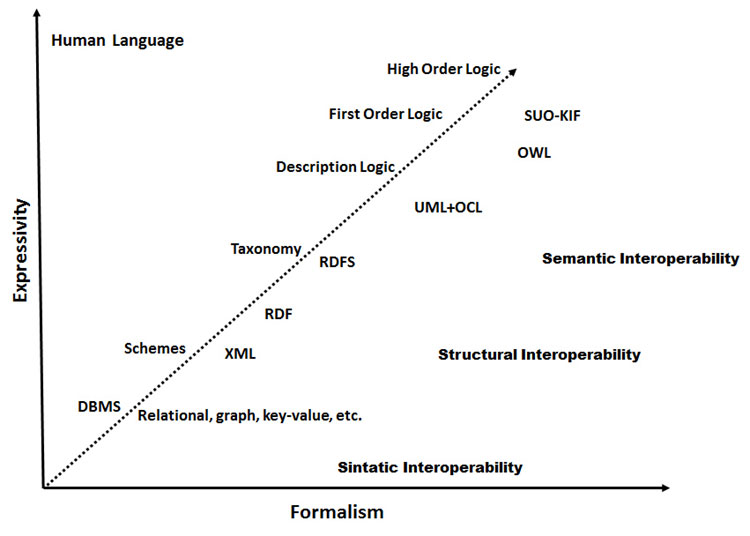}
\caption{Languages to represent knowledge and its expressivity. Source: \cite{Pease:2011a}}
\label{fig:expressiveness}
\end{figure}

The IEs, after the interconnection, must maintain the process of communication and cooperation between them. This is what represents \textbf{semantic interoperability}. The exchange of information between them begins to produce data with understandable meaning, whose interpretation is appropriate to produce the expected results, throughout the process. Such data, in appropriate repositories, are accompanied by a special formatting called \textbf{ontology}. Ontology is the resource used to represent knowledge. The ontology, that is, the knowledge associated with data in the pure state, has adequate languages to serve its purposes. Such languages vary depending on their ability to clearly express desired knowledge. Although details are discussed later in this text, the Figure \ref{fig:expressiveness} displays the main languages available in a 
comparison of its formalism and its ability to express knowledge. Note that the most powerful of languages is the natural language, which has such a restriction of interpretation by non-human procedures, which will be avoided in this project

The interoperability of A2RD IEs will be done through blockchains, one for each model implemented as we can see in \cite{Braga:2018}.

\section{Intelligent Agents}

In the early 1990s, Michael Wooldridge from his doctoral thesis in 1992 made a considerable effort in the formalization of intelligent agents \cite{Wooldridge:2014, Wooldridge:1992}. Also he established the theoretical and practical concepts involving intelligent agents, their architecture, mathematical models, logic and semantics, as well as considerations about the issues surrounding software projects and development, with approaches on the multi-disciplinarity around of intelligent agents \cite{Wooldridge:1995}. This same work recalls that a cluster of intelligent agents form a \textbf{agency}, characterization perfectly adequate to the set of IEs, however, insufficient. An agency actually consists of \textbf{muli-agents}. On two occasions, Wooldridge consolidated the ideas on multiagents \cite{Wooldridge:2002, Wooldridge:2009}. In the second edition, using an article he defines, in free translation \cite{Wooldridge:1995}:

\begin{quote}
 ``An agent is a computer system that is situated in some environment, and that is capable of autonomous action in this environment in order to meet its delegated objectives'' \cite[Chapter 2]{Wooldridge:2009}.
\end{quote}

\begin{figure}[!ht]
\centering
\includegraphics[width=1\columnwidth]{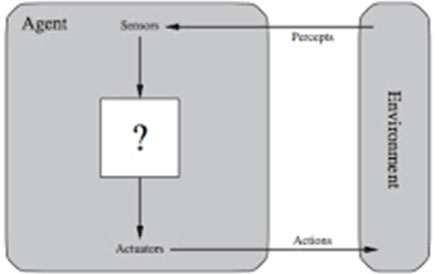}
\caption{How the agents interact with the environment. Source: \cite[página 35]{Russel:2010}}.
\label{fig:AgenteRussel}
\end{figure}

On the other hand, \cite{Russel:2010}, more simply, propose a figure to illustrate the concept of the relationship between the agent and the environment in which it is associated (Figure \ref{fig:AgenteRussel}). For them, \textit{an agent is something that has \textbf{perception} of its environment through \textbf{sensors} and acts on the environment through \textbf{actuators} '}.

\cite{Russel:2010, Wooldridge:2002, Wooldridge:2009} are appropriate texts for the development of the proposed work. The first, with its broad description of Artificial Intelligence techniques and the second a specific text about multi-agents. To them, join a book, also, essential and complementary \cite{Wooldridge:2000a}. In addition, one can not forget the text of \cite{Weiss:1999}. Everyone at the end of each chapter makes appropriate and convincing approaches to the issues discussed above, including a literature review. Naturally, \cite{Russel:2010} present a more up-to-date text.

\section{Multi-Agents}

\begin{figure}[!ht]
\centering
\includegraphics[width=1\columnwidth]{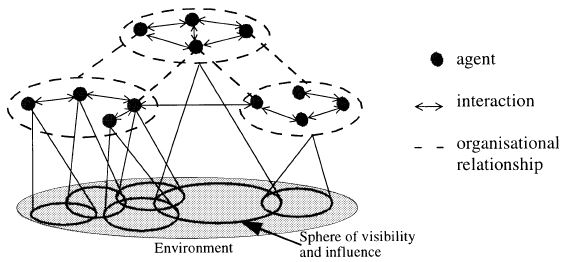}
\caption{Canonical view of an agent-based system. Source: \cite[page 281]{Jennings:2000}}
\label{fig:multiagentescanonico}
\end{figure}

In addition to the definitions of agents, in the previous sections, a very clear multi-agent model proposed by \cite{Jennings:2000} is represented in Figure \ref{fig:multiagentescanonico}. This model is suitable to establish the main motivation of the proposal of the A2RD model. Before proceeding with this analysis of the peculiarities and specific properties of an IE, it is worth noting the definition given by \cite{Jennings:2000}:

\begin{quote}
An agent is an encapsulated computer system that is situated in some environment and that is capable of flexible, autonomous action in that environment in order to meet its design objectives.
\end{quote}

Immediately, the same author clarifies some points associated with the definition of his canonical model. According to him, agents are:

\begin{enumerate}[label=(\alph*), topsep=0pt, partopsep=0pt, itemsep=0pt, parsep=0pt]
\item \textit{clearly identifiable}: as entities that solve problems, with well-defined interfaces and boundaries;
\item\textit{encapsulated in a particular environment}: they receive inputs related to the state of their environment, through sensors and act on the environment through actuators (which he calls \textit{effectors}); 
\item \textit{designed to meet a specific goal}: they have particular goals to meet;
\item \textit{autonomous}: they have control over both their internal state and their own behavior. The control characteristic about its own behavior is what distinguishes agents from objects;
\item \textit{able to exhibit flexible problem solver behavior}: in addressing their goals they need to be both reactive (able to respond in time to changes that occur in their environment) and proactive (empowered to act ahead of your future goals). 
\end{enumerate}

These observations, complemented by \cite{Wooldridge:1995, Wooldridge:1997} make evident the definitions of multi-agents that continues a very active and effervescent area of research but with many aspects still to be debated. Thus, the analysis of the figure \ref{fig:multiagentescanonico} can proceed, on which two aspects stand out:

\begin{enumerate}[label=(\roman*), topsep=0pt, partopsep=0pt, itemsep=0pt, parsep=0pt]
\item There is not a complete mesh of \textit{full mesh}. Some agents do not communicate directly with other agents. \label{label:mesh}
\item In relation to the environment, the agents act in specific sub-domains and more than one agent can act on the same sub-domain.  
\end{enumerate}

\section{Communication between Intelligent Agents}

\begin{figure}[!ht]
\centering
\includegraphics[width=1\columnwidth]{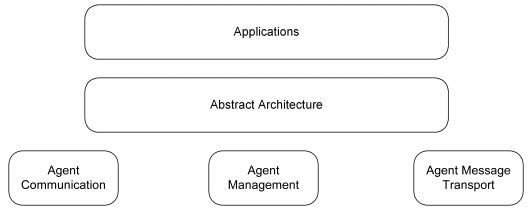}
\caption{FIPA specification category groups.}
\label{fig:fipa}
\end{figure}

There is a huge research effort in the direction of defining the communication properties between agents. One such initiative is the \textit{Foundation for Intelligent Physical Agents}\footnote{\url{http://www.fipa.org}} (FIPA). founded in 1995 with the aim of developing standards aimed at systems of agents. These patterns, specified in categories are grouped\footnote{\url{http://www.fipa.org/repository/bysubject.html}} according to Figure \ref{fig:fipa}.

FIPA standards systematically use Software Engineering and, in particular, Unified Modeling Language (UML), as meta-language cite {Eriksson2004, Eriksson: 2000}. In fact, FIPA recommends extensions for UML dealing with agents. \cite{Bernon:2005} present a comparison of several meta-models used in some known methodologies, in search of common aspects between them. In addition, in a UML approach, \cite{Bauer:2001} provides a very expressive overview and presents an extension of the UML, called Agent Unified Modeling Language (AUML). \cite{Odell:2001} describe the UML meeting the specification \cite{FIPA00036}. In \cite{Bauer:2000}, the AUML extension is described comprehensively. Object Management Group (OMG) discusses agents\footnote{\url{http://www.objs.com/agent/}}, and recommends a specific environment for AUML\footnote{\url{http://www.auml.org/}}.

The Abstract Architecture \cite{FIPA00001} is the prerequisite for moving on to other specifications. In \cite{FIPA00023}, the proposal for an inter-agent communication language (ACL) that gave rise to Java Agent Development Framework (JADE), whose best-known original document is \cite{Bellifemine:1999} followed by a complementary article \cite{Bellifemine:2008} and a much more complete text in \cite{Bellifemine:2007}. The importance of the environment, in which the agents interact, is characterized in a very lucid manner in \cite{Odell:2003}.

All active FIPA specifications are listed in Table \ref{tab:FIPA}. The specifications can be classified according to the index indicated in the column of Code: with ($^1$) are those relating to Applications (\textit{Applications}), with ($^4$), associated ($^5$) those associated with the transport of messages between agents (\textit{Agent Message Transport}), with ($^6$) those associated with the repository of messages, and ($^*$) refer to the group related to the specifications of the interaction protocol (\textit{Interaction Protoccol}), between agents.

\setcounter{rowno}{0}
\begin{table}[ht!]
\tiny
\caption{FIPA active recommendations}
\centering
\begin{tabular}{|>{\stepcounter{rowno}\therowno.} l | p{1cm}|p{1.5cm}|p{3cm}| p{0.5cm}|}
\hline  \hline 
 \multicolumn{1}{|c|}{\textbf{\#}} & \multicolumn{1}{| c |}{\textbf{Código}} & \multicolumn{1}{| c |}{\textbf{Situação}} & \multicolumn{1}{| c |}{\textbf{Descrição}}  & \multicolumn{1}{| c |}{\textbf{Ref}}\\ 
 \hline
   & FIPA00001$^4$  & Standard & Abstract Architecture Specification & \cite{FIPA00001}\\ 
\hline
   & FIPA00008$^4$  & Standard & Content Language Specification & \cite{FIPA00008} \\ 
\hline
  & FIPA00009$^4$  &  Experimental & CCL Content Language Specification & \cite{FIPA00009} \\ 
\hline
  & FIPA00010$^4$  & Experimental & KIF Content Language Specification & \cite{FIPA00010} \\ 
\hline
  & FIPA00011$^4$  & Experimental & RDF Content Language Specification& \cite{FIPA00011} \\ 
\hline
  & FIPA00014$^1$ & Standard & Nomadic Application Support Specificationn & \cite{FIPA00014} \\ 
\hline
  & FIPA00023$^4$ & Standard  & Agent Management Specification & \cite{FIPA00023} \\ 
\hline
  & FIPA00026$^{4,*}$  & Standard  & Request Interaction Protocol Specification & \cite{FIPA00026} \\ 
\hline
  & FIPA00027$^{4,*}$  & Standard  & Query Interaction Protocol Specification & \cite{FIPA00027} \\ 
\hline
  & FIPA00028$^{4,*}$  & Standard  & Request When Interaction Protocol Specification & \cite{FIPA00028} \\ 
\hline
  & FIPA00029$^{4,*}$  & Standard  & Contract Net Interaction Protocol Specification & \cite{FIPA00029} \\ 
\hline
  & FIPA00030$^{4,*}$ & Standard  & Iterated Contract Net Interaction Protocol Specification & \cite{FIPA00030} \\ 
\hline
  & FIPA00031$^{4,*}$ & Standard  & English Auction Interaction Protocol Specification & \cite{FIPA00031} \\ 
\hline
  & FIPA00032$^{4,*}$ & Standard  & Dutch Auction Interaction Protocol Specification & \cite{FIPA00032} \\ 
\hline
  & FIPA00033$^{4,*}$ & Standard  & Brokering Interaction Protocol Specification & \cite{FIPA00033} \\ 
\hline
  & FIPA00034$^{4,*}$ & Standard  & Recruiting Interaction Protocol Specification & \cite{FIPA00034} \\ 
\hline
  & FIPA00035$^{4,*}$ & Standard  & Subscribe Interaction Protocol Specification & \cite{FIPA00035} \\ 
\hline
  & FIPA00036$^{4,*}$ & Standard  & Propose Interaction Protocol Specification & \cite{FIPA00036} \\ 
\hline
  & FIPA00037$^4$ & Standard  & Communicative Act Library Specification & \cite{FIPA00037} \\ 
\hline
  & FIPA00061$^4$ & Standard  & ACL Message Structure Specification & \cite{FIPA00061} \\ 
\hline
  & FIPA00067$^4$ & Standard  & Agent Message Transport Service Specification & \cite{FIPA00067} \\ 
\hline
  & FIPA00069$^4$ & Standard  & ACL Message Representation in Bit-Efficient Specification & \cite{FIPA00069} \\ 
\hline
  & FIPA00070$^4$ & Standard  & ACL Message Representation in String Specification & \cite{FIPA00070} \\ 
\hline
  & FIPA00071$^4$ & Standard  & ACL Message Representation in XML Specification & \cite{FIPA00071} \\ 
\hline
  & FIPA00075$^{4,5}$ & Standard  & Agent Message Transport Protocol for IIOP Specification & \cite{FIPA00075} \\ 
\hline
  & FIPA00076$^{4,5}$ & Standard  & Agent Message Transport Protocol for WAP Specification & \cite{FIPA00076} \\ 
\hline
  & FIPA00079$^{1,4}$ & Standard  & Agent Software Integration Specification & \cite{FIPA00079} \\ 
\hline
  & FIPA00080$^1$ & Standard  & Personal Travel Assistance Specification & \cite{FIPA00080} \\ 
\hline
  & FIPA00081$^1$ & Standard  & Audio-Visual Entertainment and Broadcasting Specification & \cite{FIPA00081} \\ 
\hline
  & FIPA00082$^1$ & Experimental  & Network Management and Provisioning Specification & \cite{FIPA00082} \\ 
\hline
  & FIPA00083$^1$ & Experimental  &  Personal Assistant Specification & \cite{FIPA00083} \\ 
\hline
  & FIPA00084$^{4,5}$ & Standard  & Agent Message Transport Protocol for HTTP Specification & \cite{FIPA00084} \\ 
\hline
  & FIPA00085$^{4,6}$ & Standard  & Agent Message Transport Envelope Representation in XML & \cite{FIPA00085} \\ 
\hline
  & FIPA00086$^4$ & Experimental  & Ontology Service Specification & \cite{FIPA00086} \\ 
\hline
  & FIPA00088$^{4,6}$ & Standard  & Agent Message Transport Envelope Representation in Bit Efficient & \cite{FIPA00088} \\ 
\hline
  & FIPA00089$^4$ & Preliminary  & Domains and Policies Specification & \cite{FIPA00089} \\ 
\hline
  & FIPA00091 & Standard  & Device Ontology Specification & \cite{FIPA00091} \\ 
\hline
  & FIPA00092$^{1,4}$ & Experimental  & Message Buffering Service Specification & \cite{FIPA00092} \\ 
\hline
  & FIPA00093$^4$ & Experimental  & Messaging Interoperability Service Specification & \cite{FIPA00093} \\ 
\hline
  & FIPA00094$^{1,4}$ & Standard  & Quality of Service Specification & \cite{FIPA00094} \\ 
\hline
  & FIPA00095$^4$ & Standard  & Agent Discovery Service Specification & \cite{FIPA00095} \\ 
\hline
  & FIPA00096$^4$ & Standard  & JXTA Discovery Middleware Specification & \cite{FIPA00096} \\ 
\hline
  & FIPA00097$^1$ & Standard  & Design Process Documentation Template & \cite{FIPA00097} \\ 
\hline
\end{tabular}
\label{tab:FIPA}
\end{table}

Although the FIPA proposal is not the only one, whether it is an attempt to standardize or not, for example, in cases of agents in \textit{peer-to-peer} environments, described in \cite{Moro:2005}, the This project deviates considerably from FIPA's proposal, making the desired scenario for the IES Agency more flexible and closer to the characteristics of the Internet Infrastructure if this concept is necessary.

\section{Related works}

Table \ref{tab:relatedworks} presents the main works, which include models that were considered during the development of A2RD.

\begin{table}[!ht]
\tiny
\centering
\caption{Related works}
\begin{tabular}{|p{1.5cm}|p{1.2cm}|p{1.1cm}| p{1cm}|p{1.7cm}|}
\hline \hline
\multicolumn{1}{|c|}{\textbf{Characteristic}} & \multicolumn{1}{|c|}{\textbf{MAPE-k}} & \multicolumn{1}{|c|}{\textbf{Others}} & \multicolumn{1}{|c|}{\textbf{Schmid}} & \multicolumn{1}{|c|}{\textbf{ANIMA}} \\ 
\hline \hline
Reference & IBM \cite{Horn:2001} & \cite{Movahedi:2012} & \cite{Schmid:2006} & \cite{Behringer:2014} (IRTF + IETF) \\ \hline
Domain & Application. IBM Products. & Application. Autonomic Architecture & Networks & Networks: autonomic nodes with the same intention \\ \hline
Integration between elements & Through an executor & Follow the MAPE-K: executor equivalent & Variation of MAPE-K & If necessary, use the Feedback Cycle \\ \hline
ID & Undefined & Undefined & Undefined & IPv6 (host interface \\ \hline
Human Interference &  High Level Objective & High Level Objective & High Level Objective. Deterministic Behavior & Intention. Autonomic Control Plane \\ \hline
Specific & Self management & Self management & Self management and self-adaptation & Self-management, network knowledge, self knowledge (self-awareness)  \\ \hline
Architecture &  Centralized, restricted scalability & Hierarchical, peer, restricted scalabilit &  Similar to FIPA's proposal: Agency & Scalable in the domain \\ \hline
\hline
\end{tabular} 
\label{tab:relatedworks}
\end{table}

The origin of the associated ideas arose from the proposal of Autonomic Computing elaborated by Paul Horn in \cite{Horn:2001}. \cite{Movahedi:2012} display details of Horn's MAPE-K model with their respective control cycles and compare several other proposals of autonomic architectures with emphasis on networks. \cite{Schmid:2006} proposes changes in the MAPE-K model, simplifying it, for elements of autonomic networks. \cite{Behringer:2014} started in the Internet Research Task Force (IRTF) NMRG group, proposals that continued in the IETF ANIMA group, with studies that are in full activity, with specific recommendations for protocols, indicating the most recent and active studies on autonomic networks.From the models described in Table \ref{tab:relatedworks}, only ANIMA proposes an identification (ID) for its autonomic functions, associating them with an IPv6 address, indicating that the autonomic functions are aggregated to the host interface.

\section{Thanks}

From Juliao Braga: Supported by CAPES – Brazilian Federal Agency for Support and Evaluation of Graduate Education within the Brazil’s Ministry of Education and was also supported by national funds through Fundação para a Ciência e a Tecnologia (FCT) with reference UID/CEC/50021/2013.

\bibliographystyle{IEEEtranN}
\bibliography{pdbib}

\end{document}